\documentclass[letterpaper, 10 pt, conference]{ieeeconf}  

\IEEEoverridecommandlockouts                              

\usepackage{graphicx} 
\usepackage{amsmath} 
\usepackage{amssymb}  
\usepackage{cite}
\usepackage{stfloats}
\usepackage{bm}
\usepackage{url}
\usepackage{booktabs}
\usepackage{multirow}
\usepackage{stfloats}
\usepackage{balance}
\usepackage{hyperref}
\hypersetup{hidelinks}
\usepackage{float}

\title{\LARGE \bf
Controllable and Photorealistic Pedestrian Risky Motion Generation for End-to-End Driving Safety Evaluation}

\author{Siyuan Liu$^{1,2}$, Miao Li$^{2}$, Haibao Yu$^{2}$, Haohong Lin$^{2}$, Qing Zhou$^{1}$, Bingbing Nie$^{1,3,\dagger}$ and Ding Zhao$^{2}$ \\
Project Page: \url{https://controlped.netlify.app}%
\thanks{$^{\dagger}$Correspondence to Bingbing Nie (nbb@tsinghua.edu.cn)}%
\thanks{$^{1}$Siyuan Liu, Qing Zhou and Bingbing Nie are with School of Vehicle and Mobility, Tsinghua University. (email: siyuan-l22@mails.tsinghua.edu.cn; zhouqing@tsinghua.edu.cn; nbb@tsinghua.edu.cn)}%
\thanks{$^{2}$Siyuan Liu, Miao Li, Haibao Yu, Haohong Lin and Ding Zhao are with Carnegie Mellon University. (email:siyuanl5@andrew.cmu.edu; limiao@andrew.cmu.edu; yuhaibao94@gmail.com; haohongl@andrew.cmu.edu; dingzhao@andrew.cmu.edu)}%
\thanks{$^{3}$Bingbing Nie is also with State Key Laboratory of Intelligent Green Vehicle and Mobility, Tsinghua University. (email: nbb@tsinghua.edu.cn)}%
}

\begin{document}

\maketitle
\thispagestyle{empty}
\pagestyle{empty}

\begin{abstract}

Evaluating end-to-end autonomous driving under rare, safety-critical vehicle-pedestrian interactions requires photorealistic, sensor-level scenarios. However, trajectory-based scenario generators cannot synthesize raw visual observations, whereas video-based approaches lack controllability. To bridge this gap, we present ControlPed, a novel framework that combines trajectory-level conflict synthesis with 3D Gaussian Splatting (3DGS) to generate photorealistic, motion-controllable safety-critical scenarios. Built upon HazardPed, a dataset derived from 10,352 traffic videos comprising 422 conflict trajectories, HD maps, and 857 annotated 3D human motions, ControlPed first generates conflict trajectories, lifts them into 3D human motion sequences via text-conditioned motion diffusion, and finally renders multi-view sensor observations using animatable 3DGS avatars. Safety evaluation in 88 rendered photorealistic scenarios reveals that seven leading end-to-end driving models suffer a severe performance drop, with their mean HDScore plunging from 88.8 to 47.4, exposing major failure modes under dangerous pedestrian behaviors. The dataset and testing benchmarks will be released to facilitate safety assessment of vehicle-pedestrian interactions.

\end{abstract}

\section{INTRODUCTION}

Evaluating end-to-end autonomous driving requires sensor-level safety-critical data, since raw observations are directly mapped to future trajectories~\cite{hu2023planning}. However, high-risk vehicle-pedestrian interactions are rare in real-world driving logs. This motivates synthesizing vehicle-pedestrian conflict scenarios at the sensor-level to evaluate models under realistic deployment conditions~\cite{zhou2025hugsim}.

Existing approaches for safety-critical scenario generation follow trajectory-based or video-based paradigms. For trajectory-based methods, adversarial~\cite{liu2026adv} methods often generate unnatural trajectories, and learning-based~\cite{lin2025causal} methods suffer from data scarcity. Both of them lack multi-view video rendering. In contrast, video-based methods satisfy sensor-level requirements~\cite{liu2026safegen,fu2025controllable}, but lack behavioral controllability. Recent technical advances provide new ways to address these challenges. 3DGS converts driving logs into photorealistic scenes~\cite{chen2025omnire}, text-conditioned motion generation models enable fine-grained control~\cite{tevet2022human}, and animatable avatars facilitate dynamic integration~\cite{lei2024gart}. However, because existing motion datasets focus on daily tasks~\cite{punnakkal2021babel}, models trained on them fail to capture safety-critical pedestrian dynamics~\cite{huang2024stablemofusion}. Bridging this gap requires constructing safety-critical pedestrian datasets to train generative models for controllable, photorealistic scene editing and end-to-end algorithms evaluation.

\begin{figure}[t]
    \vspace{+5pt}
    \centering
    \includegraphics[width=\columnwidth]{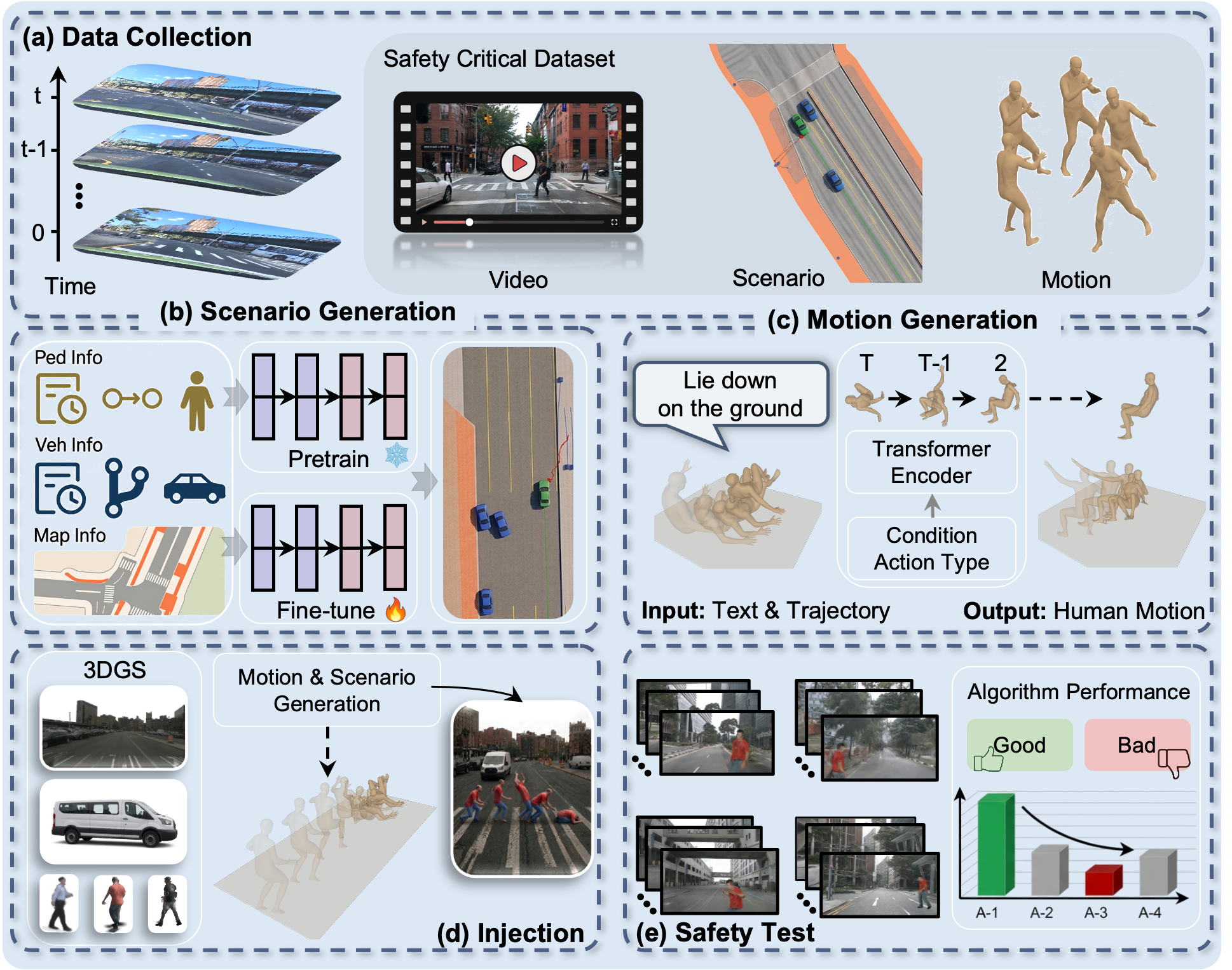}
    \caption{ControlPed framework. (a) Data collection: filtering risky cases from traffic videos to extract conflict trajectories, HD maps, and pedestrian motions. (b) Scenario generation: synthesizing vehicle-pedestrian conflict trajectories. (c) Motion generation: lifting 2D trajectories into text-conditioned 3D human motion sequences. (d) Hazard injection: compositing dynamic pedestrians into 3DGS reconstructed scenes. (e) Safety test: evaluating end-to-end driving algorithms on rendered multi-view scenarios.}
    \label{fig:framework}
    \vspace{-15pt}
\end{figure}


Our key insight is that safety-critical scenario generation requires hierarchical decomposition: macro-level collision paths governed by trajectory control, micro-level actions guided by joint dynamics, and sensor-level inputs realized via photorealistic rendering. To this end, we propose ControlPed, a three-stage framework comprising trajectory generation, motion synthesis, and photorealistic rendering. Trained on 422 real conflict trajectories and 857 hazardous human motion sequences (Fig.~\ref{fig:framework}a), ControlPed couples a conflict trajectory generator with a text-driven motion diffusion model (Fig.~\ref{fig:framework}b–c). A hard root constraint guarantees strict alignment between generated motions and target trajectories. Animatable avatars are subsequently composited into 3DGS scenes to render photorealistic camera videos (Fig.~\ref{fig:framework}d). Finally, these synthesized scenarios serve to evaluate seven end-to-end driving models, closing the loop from conflict synthesis to safety assessment (Fig.~\ref{fig:framework}e).


This work makes the following key contributions:
\begin{itemize}
    \item \textbf{HazardPed Dataset:} We present \textit{HazardPed}, a safety-critical dataset that combines real conflict trajectories, HD maps, and human motions. It bridges the scarcity of risky interaction data, providing a prior for controllable pedestrian scenario generation.
    \item \textbf{ControlPed Generator:} We propose \textit{ControlPed}, a controllable scenario synthesis method that couples trajectory-level conflict generation with text-conditioned motion diffusion. It achieves sensor-level photorealism without sacrificing fine-grained pedestrian control.
    \item \textbf{Safety Evaluation Benchmark:} We establish a photorealistic benchmark to evaluate autonomous driving safety. This benchmark enables third-party testing agencies and regulatory bodies to assess algorithm reliability before real-world deployment.
\end{itemize}

\section{RELATED WORK}

\begin{figure*}[b]
    \vspace{-15pt}
    \centering
    \includegraphics[width=\textwidth]{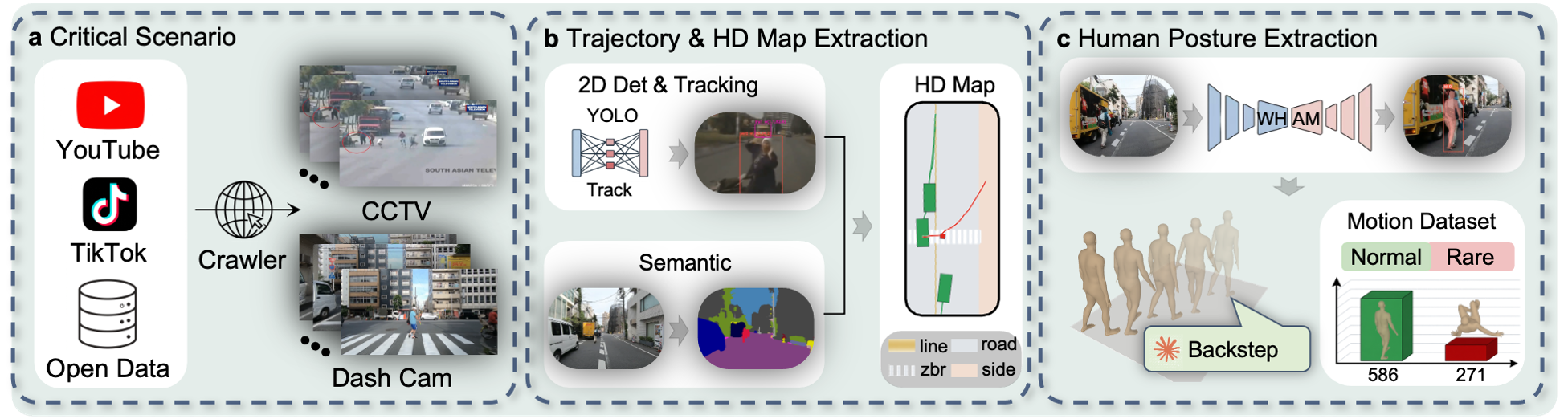}
    \vspace{-15pt}
    \caption{Dataset construction. (a) Critical scenario: collecting safety-critical clips from public datasets. (b) Trajectory and map extraction: detecting and tracking agent trajectories, and aligning them with HD maps. (c) Motion annotation: recovering 3D SMPL poses to build a text-conditioned motion dataset.
    }
    \label{fig:dataset}
\end{figure*}

\subsection{Safety-Critical Scenario Generation}

Trajectory-based scenario generators synthesize hazardous interactions via behavior perturbation~\cite{liu2026adv,jing2026counterscene} or traffic priors~\cite{lin2025causal}. However, their 2D bird's-eye-view representations strip away key pedestrian dynamics essential for vision-based end-to-end models. Conversely, video-based approaches introduce sensor-level realism via asset insertion or pose-guided editing~\cite{liu2026safegen, fu2025controllable}, yet lack fine-grained trajectory or human control. Asset insertion struggles to enforce precise terminal collision states, while 2D pose editing lacks metric grounding, preventing control over real-world vehicle–pedestrian distances. Consequently, achieving joint controllability over macro trajectories and micro pedestrian motions in photorealistic rendering remains an open challenge.

\subsection{Human Motion Generation and Datasets}

Text-conditioned human motion generation models, built on diffusion architectures~\cite{tevet2022human, huang2024stablemofusion}, synthesize diverse motions from language but lack spatial grounding to specify coordinate dynamics. Although recent spatial control techniques incorporate trajectory or sparse joint constraints~\cite{cohan2024flexible,xie2024omnicontrol,shafir2024human}, their applicability remains bounded by the underlying training data. Standard motion benchmarks built on AMASS, HumanML3D, and BABEL focus almost exclusively on daily activities~\cite{zhu2023human}. Consequently, critical conflict dynamics, such as abrupt dodging or freezing under imminent vehicle threat, remain remarkably scarce, hindering the generation of safety-critical scenarios.

\subsection{Scenario Reconstruction and End-to-End Evaluation}

3DGS enables photorealistic rendering and has been widely adopted in driving scenario reconstruction~\cite{zhou2024drivinggaussian}. To handle dynamic objects, recent approaches decouple foreground actors from background modeling, allowing moving entities to be controlled independently~\cite{chen2025omnire}. Advances in pose-driven animatable avatars further free pedestrians from rigid log replaying, enabling the insertion of hazardous human behaviors into reconstructed scenes. For safety evaluation, existing paradigms face clear trade-offs: open-loop tests on passive logs lack reactivity, simulators exhibit a pronounced sim-to-real gap. Recently, reconstruction-based closed-loop simulations combine realism with interactivity by evaluating models on rendered real-world scenes~\cite{zhou2025hugsim}, yet they lack critical pedestrian-involved hazards. Therefore, injecting controllable pedestrian behaviors into photorealistic reconstructed scenes is essential for end-to-end safety evaluation.

\section{HAZARDPED DATASET}

To enable controllable injection of hazardous pedestrian motions and resolve the scarcity of safety-critical interaction data, we construct HazardPed by filtering dynamic interactions from web videos and open-source datasets. As illustrated in Fig.~\ref{fig:dataset}, HazardPed integrates diverse vehicle-pedestrian conflict trajectories, aligned HD maps, and paired text-conditioned human motion data.

\subsection{Safety-critical scenario dataset} 

We collected 10,352 real-world traffic videos from YouTube and public datasets~\cite{kotseruba2016joint}. Following the PedGen pipeline, we performed 2D pedestrian tracking, camera pose estimation, and SMPL body fitting, retaining 3,497 valid clips after post-processing~\cite{liu2025learning}. Object trajectories were extracted using YOLO11 and BoT-SORT~\cite{khanam2024yolov11,aharon2022bot}. To map 2D detections into global world coordinates, homography matrices for static CCTV feeds were calibrated based on average human height, while UniDepthV2 and DROID-SLAM estimated camera intrinsics and ego-vehicle poses for dashcam feeds~\cite{liu2026monocular}. Finally, Mask2Former~\cite{cheng2022masked} extracted road semantics, yielding 422 scenarios with 566 vehicle-pedestrian conflict pairs.

\begin{figure*}[b]
    \vspace{-15pt}
    \centering
    \includegraphics[width=\textwidth]{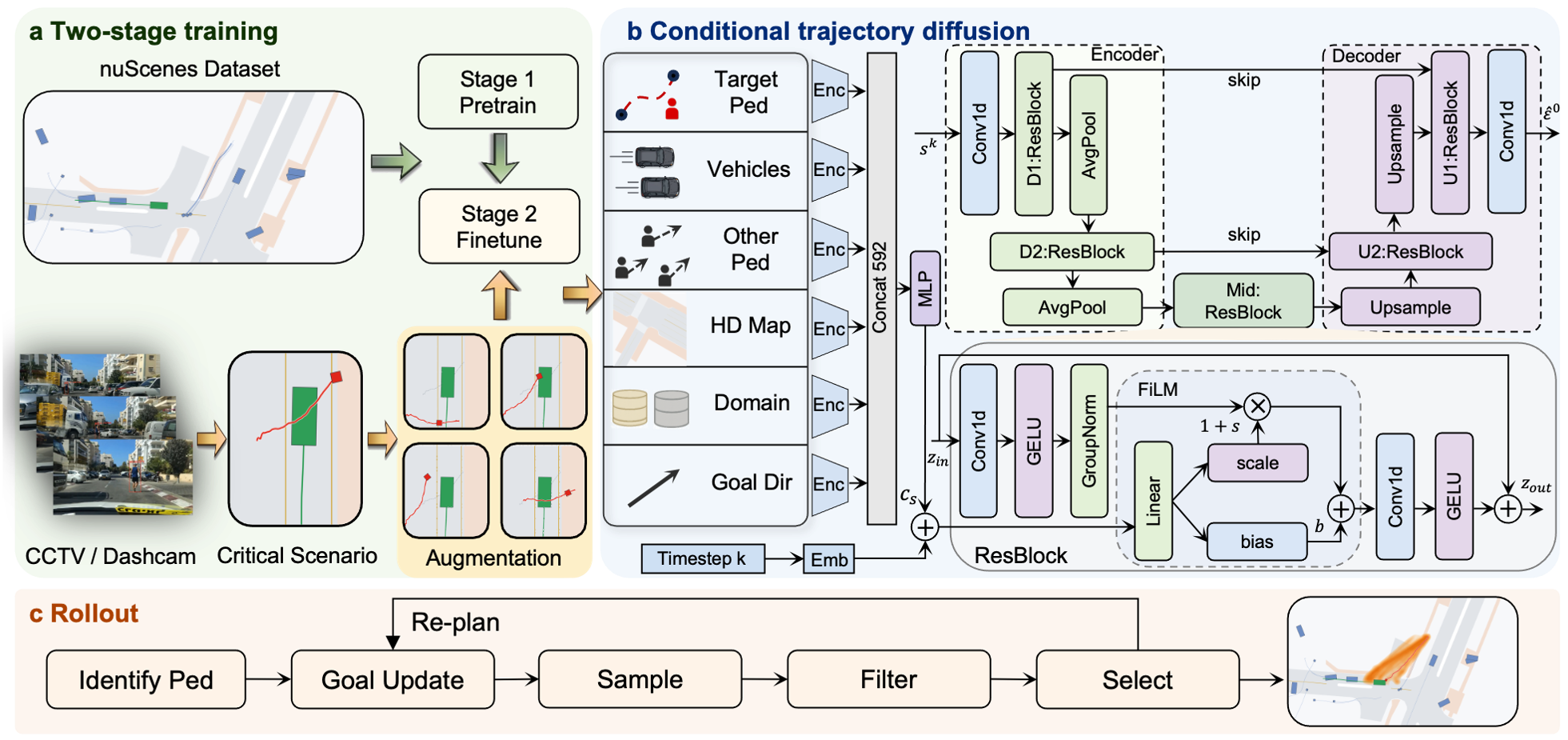}
    \vspace{-15pt}
    \caption{Hazardous trajectory generation. (a) Two-stage training: pretraining on nuScenes, followed by fine-tuning on conflict data. (b) Conditional trajectory diffusion: encoding heterogeneous context into a unified condition vector $\bm{c}_s$, which modulates a 1D U-Net denoiser via FiLM blocks. (c) Rollout: the conflict controller updates the goal heading $\bm{g}$ every $0.4$\,s to sample and filter candidate trajectories.}
    \label{fig:trajectory}
\end{figure*}

\subsection{Human Motion Extraction and Annotation} 

Following human verification, 1,473 valid motion clips were retained. We adopted the vocabulary format of BABEL~\cite{punnakkal2021babel} for action annotation. Human motions were rendered into video frames, allowing Claude to generate initial textual descriptions and action categories. Precise temporal boundaries for action segments were then established using frame-wise kinematic signals. Upon final human review, we obtained 586 navigation motion clips and 271 hazardous reaction motion clips. Each annotated sample consists of a pedestrian motion sequence paired with textual annotations, denoted as $(\bm{x}, a, y)$, where $\bm{x} = \{(\bm{t}_\tau, \bm{\phi}_\tau, \bm{\theta}_\tau)\}_{\tau=1}^{T}$ across $T$ frames. Here, $\bm{t}_\tau \in \mathbb{R}^3$ and $\bm{\phi}_\tau \in SO(3)$ specify root translation and orientation, while $\bm{\theta}_\tau \in SO(3)^{21}$ captures the rotations of 21 body joints. The scalar $a \in \mathcal{A}$ indicates the action category, and $y$ represents the natural language description. To protect privacy, we release only processed trajectories, HD maps, and motion data, excluding raw source videos. The dataset is available at \url{https://tinyurl.com/ControlPed}.

\section{CONTROLPED: PEDESTRIAN GENERATION}

\subsection{Hazardous Trajectory Generation}

\textbf{Task Definition.} We define hazardous pedestrian trajectory generation in a pedestrian-centric frame. Given an observed historical trajectory $\bm{o} = \{\bm{p}_\tau\}_{\tau=-H+1}^{0}$, our goal is to synthesize the future trajectory $\bm{s} = \{\bm{p}_\tau\}_{\tau=1}^{N}$ conditioned on the surrounding scenario and the goal heading, modeled as $\bm{s} \sim p(\bm{s} \mid \bm{o}, \mathcal{V}, \mathcal{P}, \mathcal{M}, b, \bm{g})$. Here, $\mathcal{V}$ and $\mathcal{P}$ represent surrounding vehicles and other pedestrians, while $\mathcal{M} \in \mathbb{R}^{4 \times G \times G}$ encodes local map semantics. The indicator $b \in \{0, 1\}$ specifies whether the scene originates from nuScenes~\cite{caesar2020nuscenes} or HazardPed. Finally, the goal heading $\bm{g} \in \mathbb{R}^2$ ($\|\bm{g}\| = 1$) is a unit vector specifying the pedestrian's intended direction of motion.

\textbf{Conditional Trajectory Diffusion.} Our trajectory generator follows the conditional diffusion paradigm. Given a pedestrian trajectory $\bm{s}$, the forward process defines a Markov noising chain $\{\bm{s}^k\}_{k=0}^{K}$ over $K$ steps. A denoising network $\bm{\epsilon}_\omega(\bm{s}^k, k, \bm{c}_s)$ is trained to predict the added noise given the noisy state $\bm{s}^k$, diffusion step $k$, and condition vector $\bm{c}_s$. As illustrated in Fig.~\ref{fig:trajectory}b, dedicated encoders process heterogeneous inputs to construct $\bm{c}_s$. Object sets $\mathcal{V}$ and $\mathcal{P}$ are processed via temporal convolutions with masked mean pooling, local map $\mathcal{M}$ is encoded using strided convolutions with global average pooling, while $\bm{o}$, $b$, and $\bm{g}$ are embedded through MLPs. The concatenated features are projected into $\bm{c}_s$ and added to the time embedding of $k$.

We adopt a 1D U-Net denoiser where the condition vector $\bm{c}_s$ modulates residual blocks via feature-wise linear modulation (FiLM)~\cite{perez2018film}. Given an intermediate feature map $\bm{z}$, a linear projection computes scale and shift parameters from $\bm{c}_s$ to modulate normalized activations:
\begin{equation}
\mathrm{FiLM}(\bm{z} \mid \bm{c}_s) = \big(1 + \gamma(\bm{c}_s)\big) \odot \mathrm{GN}(\bm{z}) + \beta(\bm{c}_s),
\label{eq:film}
\end{equation}
where $\mathrm{GN}(\cdot)$ denotes group normalization and $\odot$ represents the channel-wise product. Stacking these blocks yields the predicted noise $\bm{\epsilon}_\omega(\bm{s}^k, k, \bm{c}_s)$, which is optimized using a mean squared error loss:
\begin{equation}
\mathcal{L}_{\mathrm{s}} = \mathbb{E}_{k \in [1,K],\, (\bm{s}, \bm{c}_s) \in \mathcal{D}} \left[ \left\| \bm{\epsilon} - \bm{\epsilon}_\omega(\bm{s}^k, k, \bm{c}_s) \right\|_2^2 \right],
\label{eq:traj_loss}
\end{equation}
where $\mathcal{D}$ denotes the training set. For classifier-free guidance, $\bm{g}$ and $\mathcal{M}$ are randomly replaced with learnable null embeddings during training.

\textbf{Direction Augmentation and Two-Stage Training.} To overcome the limited scale of real-world conflict, we augment training data by bending future pedestrian trajectories into various directions (Fig.~\ref{fig:trajectory}a). For each scenario, historical trajectories, surrounding agents, and HD maps remain unchanged, while future positions are progressively deflected. Given a target deflection angle $\delta$, the velocity vectors $\bm{v}_\tau = \bm{p}_\tau - \bm{p}_{\tau-1}$ are rotated by $\delta\,\tau/N$ and integrated:
\begin{equation}
\bm{p}'_\tau = \bm{p}_0 + \sum_{j=1}^{\tau} \bm{Q}\!\left(\frac{j}{N}\delta\right) \bm{v}_j,
\label{eq:curving}
\end{equation}
where $\bm{Q}(\cdot)$ denotes the 2D rotation matrix and $N$ represents the future trajectory horizon. By pairing an identical historical context with varied future directions, this process expands the coverage of the goal heading $\bm{g}$, allowing trajectory synthesis when $\bm{g}$ points to oncoming vehicles.

Training proceeds in two stages using the same network architecture and loss functions, adjusting only the training datasets and learning rates. Stage-1 trains on 700 scenes from the nuScenes dataset to capture general pedestrian trajectory priors. Stage-2 finetunes the model on the augmented conflict dataset. Because nuScenes contains normal transit behaviors without conflicts, Stage-1 lacks scenarios where the goal heading $\bm{g}$ points toward a vehicle. Stage-2 compensates for these edge cases, ensuring that the trajectory generator adheres to $\bm{g}$ under safety-critical conditions.

\textbf{Conflict Controller.} To construct safety-critical scenarios, a conflict controller guides the trajectory generation model. We first filter vehicle-pedestrian pairs using reachability-based risk scores. Along the vehicle future path, candidate interception frames $\tau$ satisfying speed feasibility $u(\tau) = \|\bm{p}^{\mathrm{veh}}_\tau - \bm{p}^{\mathrm{ped}}_{0}\| / ((\tau - \tau_0)\Delta t) \leq u_{\mathrm{max}}$ are evaluated, where $\tau_0$ is the current frame index and $\Delta t$ is the time step. Candidate frames within the interception window are assigned a cost evaluating heading and velocity deviations:
\begin{equation}
\mathrm{cost}(\tau) = \frac{\Delta\psi(\tau)}{\pi} + \frac{|u(\tau) - u_0|}{u_{\mathrm{max}}},
\end{equation}
where $\Delta\psi(\tau)$ is the angle between the current heading of the pedestrian and the direction toward the interception point, and $u_0$ is the pedestrian's current speed. We apply softmax sampling to introduce variance across scenarios. The selected frame $\tau^{*}$ dictates the conflict timing, while the target gap $m^{*} \sim U[m_{\mathrm{min}}, 0]$ determines conflict severity, where $m^{*} = 0$ corresponds to a grazing contact and increasingly negative values to deeper collisions. Thus, the tuple $(\tau^{*}, m^{*})$ fully parameterizes the conflict geometry.

\begin{figure*}[b]
    \vspace{-15pt}
    \centering
    \includegraphics[width=\textwidth]{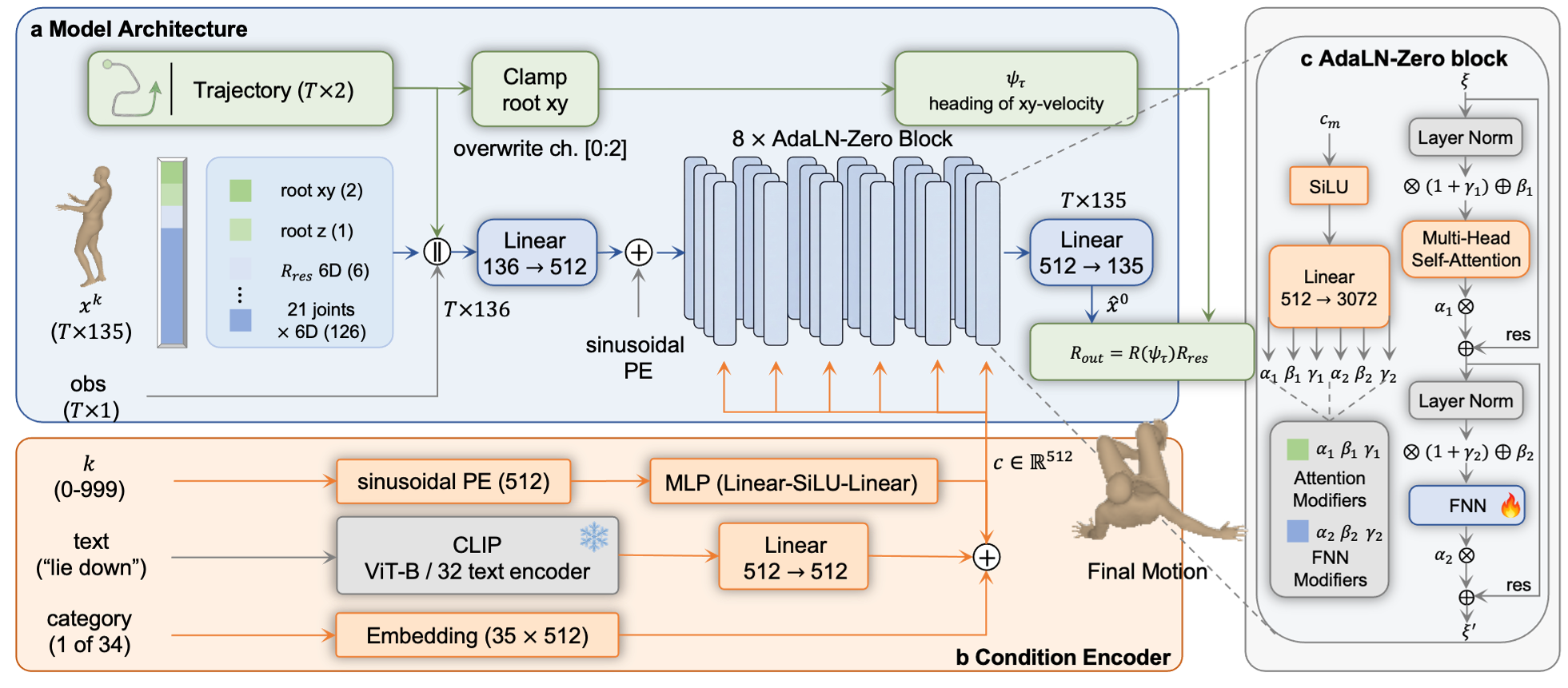}
    \vspace{-15pt}
    \caption{Human motion generation. (a) Model architecture: a diffusion denoiser lifts the ground trajectory into full-body motion. (b) Condition encoder: fusing diffusion step, text, and action category into one vector. (c) AdaLN-Zero block: the condition modulates each transformer layer.}
    \label{fig:motion}
\end{figure*}

Replanning occurs every $0.4$\,s with a temporal penalty $\eta|\tau - \tau^{*}|\Delta t$ to anchor conflict timing near $\tau^{*}$ (Fig.~\ref{fig:trajectory}c). The resulting target point is transformed into the pedestrian coordinate and normalized to form the goal heading $\bm{g}$. Candidate trajectories are sampled and filtered by pedestrian speed limits, drivable area, and threat constraints with surrounding vehicles. Therefore, the controller interfaces with the diffusion network via the goal heading $\bm{g}$.

\subsection{Human Motion Generation}

\textbf{Task Definition.} We formulate trajectory-conditioned human motion generation as lifting a 2D trajectory $\{\bm{p}_\tau\}_{\tau=1}^{T}$ into motion $\bm{x} = \{\bm{x}_\tau\}_{\tau=1}^{T}$ aligned with the input trajectory. At each frame, the motion state $\bm{x}_\tau \in \mathbb{R}^{135}$ stacks the root translation $\bm{t}_\tau$, root orientation $\bm{\phi}_\tau$ in 6D representation, and body joint rotations $\bm{\theta}_\tau$ across 21 joints in 6D. The synthesis is guided by the action category $a \in \mathcal{A}$, the corresponding text prompt $y$, and the trajectory $\{\bm{p}_\tau\}_{\tau=1}^{T}$. Rather than encoding the trajectory through an external feature extractor, we directly assign $\{\bm{p}_\tau\}_{\tau=1}^{T}$ to the first two channels of $\bm{x}_\tau$ and keep them clamped throughout generation. Thus, motion synthesis is formulated as an inpainting task where the denoiser completes the remaining 133 channels.

\textbf{Motion Diffusion with Trajectory Anchoring.} Unlike trajectory generation, our motion synthesis directly predicts clean motion $\hat{\bm{x}} = F(\bm{x}^k, k, \bm{c}_m)$ from noisy motion $\bm{x}^k$ at the diffusion step $k$ over a noising chain $\{\bm{x}^k\}_{k=0}^{K}$. During diffusion, the root horizontal channels are overwritten by the input trajectory at every step and excluded from loss computation. To factorize motion control, the root heading $\psi_\tau$ is derived by smoothing the differentiation of the trajectory, leaving the network to predict only the residual orientation:
\begin{equation}
\bm{R}_{\mathrm{res}} = \bm{R}_z(-\psi_\tau)\,\bm{R}_{\mathrm{full}},
\label{eq:reyaw}
\end{equation}
where $\bm{R}_{\mathrm{full}} \in SO(3)$ is the absolute orientation of the root and $\bm{R}_z(\cdot)$ denotes rotation about the vertical axis. At the output, the absolute orientation is recomposed as $\bm{R}_{\mathrm{full}} = \bm{R}_z(\psi_\tau)\bm{R}_{\mathrm{res}}$, ensuring that the network focuses on body turns relative to the path rather than the global direction.

The backbone is an eight-layer pre-LN transformer with a hidden dimension of 512 and 8 attention heads. The motion condition vector $\bm{c}_m \in \mathbb{R}^{512}$ sums the diffusion time embedding, the frozen CLIP text embedding of $y$, and the learned action category embedding of $a$, modulating every sublayer via adaLN-Zero~\cite{peebles2023scalable}:
\begin{equation}
\bm{\xi} \leftarrow \bm{\xi} + \alpha \odot \mathrm{SubLayer}\big((1+\gamma) \odot \mathrm{LN}(\bm{\xi}) + \beta\big),
\label{eq:adaln}
\end{equation}
where $\bm{\xi}$ is the hidden activation state, $\mathrm{LN}(\cdot)$ denotes layer normalization, and $\odot$ represents the element-wise product. Each layer regresses six modulation vectors from $\bm{c}_m$ through a zero-initialized linear layer, forming triplets $(\gamma, \beta, \alpha)$ for self-attention and feed-forward sublayers. Here, $(\gamma, \beta)$ rescales and shifts normalized activations, while $\alpha$ gates the sublayer residual output. Modulating every sublayer increases the condition at every depth, preventing the control signal from being diluted as it propagates.

The model is trained by minimizing a composite loss:
\begin{equation}
\mathcal{L}_{\mathrm{m}} = \mathcal{L}_{\mathrm{rot}} + \lambda_{v} \mathcal{L}_{v} + \lambda_{p} \mathcal{L}_{p} + \lambda_{c} \mathcal{L}_{c} + \lambda_{z} \mathcal{L}_{z} + \lambda_{s} \mathcal{L}_{s},
\label{eq:motion_loss}
\end{equation}
where $\lambda_{(\cdot)}$ are the balance weights. $\mathcal{L}_{\mathrm{rot}}$ measures the reconstruction error across the 133 generated unclamped channels, while $\mathcal{L}_{z}$ and $\mathcal{L}_{s}$ upweight root height and spine orientation channels to prevent feature dilution. The remaining terms enforce kinematic fidelity in world space via forward kinematics. $\mathcal{L}_{p}$ and $\mathcal{L}_{v}$ penalize 3D joint position and interframe velocity errors, while $\mathcal{L}_{c}$ suppresses foot sliding during ground contact phases.

During inference, denoising begins from gaussian noise with root horizontal channels clamped to the reference 2D trajectory. The text prompt $y$ and the action category $a$ are jointly dropped to enable classifier-free guidance. Long motion sequences are synthesized by conditioning each sliding window on a historical frame prefix. Navigation texts are assigned segment-wise according to the curvature of the generated hazardous trajectory, and a reaction motion sampled from the hazardous categories is concatenated at the conflict point. Linear blending is then applied across the window seams to ensure kinematic continuity.

\subsection{Pedestrian Injection and Rendering.}

We reconstruct 88 nuScenes cases using OmniRe~\cite{chen2025omnire}, representing pedestrian nodes as drivable gaussian human avatars~\cite{lei2024gart}. Generated motions are transformed from canonical local frames to world coordinates to drive newly initialized pedestrian avatars. Because generated root heights lack a shared reference datum with the reconstructed terrain, direct application induces floating or ground penetration artifacts. We mitigate this by dynamically resolving the per-frame root height $h_\tau$:
\begin{equation}
h_\tau = e(\bm{p}_\tau) - \min_{\bm{q} \in \mathcal{Q}(\bm{x}_\tau)} [\bm{q}]_z,
\label{eq:ground}
\end{equation}
where $e(\cdot)$ estimates the local ground elevation from the LiDAR points beneath the pedestrian, $\bm{p}_\tau$ is the 2D ground position, $\mathcal{Q}(\bm{x}_\tau)$ is the vertex set of the posed body mesh, and $[\cdot]_z$ takes the vertical coordinate, so that the minimum gives the offset from the root to the lowest body point. Finally, 3D gaussians from background environments, vehicles, and dynamic pedestrians are aggregated and rasterized into multi-view video streams for downstream evaluation (Sec.~\ref{sec:experiments}).

\section{EXPERIMENTS}
\label{sec:experiments}

Sec. V details the settings, baselines, and evaluation results. We evaluate ControlPed across three core dimensions: trajectory generation performance, 3D motion synthesis quality, and end-to-end model degradation in rendered scene. Qualitative visualizations are presented in Figs. 5-7.

\subsection{Experimental Setup.}
\textbf{Datasets.} Trajectory generation is evaluated on 92 vehicle-pedestrian conflict scenarios extracted from the nuScenes validation set. Motion generation is benchmarked on 857 annotated clips from HazardPed, covering 34 distinct motion categories. For end-to-end evaluation, we reconstruct conflict scenarios using OmniRe~\cite{chen2025omnire}, retaining 88 human inspected scenes rendered into multi-view videos.

\textbf{Trajectory generation metrics.}
We report six metrics across realism, criticality, and diversity.
Realism is measured via Fr\'echet Distance (FD) on local path coordinates and speed profiles. The distance $d(\tau)$ quantifies spatial separation between agents, modeling vehicles as rectangles and pedestrians as radius disks. Criticality is quantified by the collision rate (CR, where $\min_\tau d(\tau) \le 0$) and the hard braking rate (HBR). HBR is evaluated using an ego vehicle governed by an intelligent driver model, triggering when deceleration exceeds $3.0$\,m/s\textsuperscript{2}. Diversity is evaluated on hazardous subsets $\mathcal{H}_n = \{\, i : d_i < 1.0\,\mathrm{m} \,\}$ within each scenario $n$. Scenario coverage (Cov) measures the fraction of scenes containing at least one hazard. Geometric diversity ($M_{\mathrm{div}}$) averages pairwise DTW distances between hazardous future trajectories $\bm{s}_i, \bm{s}_j$, $i,j \in \mathcal{H}_n$, normalized by warping path length $|w_{ij}|$:

\begin{equation}
M_{\mathrm{div}}(n)
= \binom{|\mathcal{H}_n|}{2}^{-1}
  \sum_{i<j} \frac{\mathrm{DTW}(\bm{s}_i, \bm{s}_j)}{|w_{ij}|}.
\label{eq:mdiv}
\end{equation}
The width of the risk spectrum ($\mathrm{Span}$) captures the severity range
within a scenario:
\begin{equation}
\mathrm{Span}(n) = \max_{i} \max(d_i, 0) - \min_{i} \max(d_i, 0).
\label{eq:span}
\end{equation}

\textbf{Motion generation metrics.}
We evaluate motion generation using an evaluator trained on BABEL, reporting
Fr\'echet Inception Distance (FID), Multimodal Distance (MM-Dist), and
R-Precision (R@3)~\cite{tevet2022human}. To evaluate postural constraints, we propose
mean pelvis height error ($\mathrm{MPHE}$), the absolute error of the minimum pelvis height $h_i$ averaged over samples $\mathcal{D}_a$ of each low-pose action category $a \in \mathcal{A}_{\mathrm{low}}$:
\begin{equation}
\mathrm{MPHE}
= \frac{1}{|\mathcal{A}_{\mathrm{low}}|} \sum_{a \in \mathcal{A}_{\mathrm{low}}}
  \frac{1}{|\mathcal{D}_a|} \sum_{i \in \mathcal{D}_a}
  \left\lvert h_i^{\mathrm{gen}} - h_i^{\mathrm{GT}} \right\rvert .
\label{eq:mphe}
\end{equation}

The heading error deviation ($\mathrm{HED}$) evaluates directional alignment
during locomotion by computing the absolute difference between the generated and
ground-truth heading angles. To detect mode collapse masked by retrieval
metrics, we introduce the class separability ratio deviation ($\Delta\mathrm{CSR}$).
With intra-class scatter $S_{\mathrm{intra}}$ defined as the mean distance from
motion embeddings to class centroids and inter-class scatter $S_{\mathrm{inter}}$ as the mean pairwise distance between class centroids, we compute:
\begin{equation}
\mathrm{CSR} = \frac{S_{\mathrm{inter}}}{S_{\mathrm{intra}}}, \quad
\Delta\mathrm{CSR} = \mathrm{CSR}^{\mathrm{gen}} - \mathrm{CSR}^{\mathrm{real}},
\label{eq:dcsr}
\end{equation}
where positive values indicate over-separated categories, and negative values
reflect inter-class collapse.

\textbf{End-to-End Evaluation Metrics.}
We evaluate keyframe-planned trajectories using HUGSim~\cite{zhou2025hugsim} metrics: no collision ($\mathrm{NC}$), drivable area compliance ($\mathrm{DAC}$), time-to-collision ($\mathrm{TTC}$), comfort ($\mathrm{COM}$), route completion ($\mathrm{RC}$), and their composite $\mathrm{HDScore}$. Unlike closed-loop evaluation, we design an open-loop proxy for $\mathrm{RC}$, terminating the route at the first keyframe where a collision occurs within $0.5$\,s.

\textbf{Baselines.}
We compare trajectory generation with Adv-BMT~\cite{liu2026adv}, CCDiff~\cite{lin2025causal}, and CounterScene~\cite{jing2026counterscene}.
For motion generation, we evaluate MDM~\cite{tevet2022human},
CondMDI~\cite{cohan2024flexible}, StableMoFusion~\cite{huang2024stablemofusion}, PriorMDM~\cite{shafir2024human}, and OmniControl~\cite{xie2024omnicontrol}, all retrained in HazardPed under identical data splits. For end-to-end evaluation, UniAD~\cite{hu2023planning} and VAD~\cite{jiang2023vad} employ modular dense BEV representations, SparseDrive~\cite{sun2025sparsedrive} and SSR~\cite{li2025navigation} utilize sparse query representations, LAW~\cite{li2025enhancing} incorporates an implicit world model, DiffusionDrive~\cite{liao2025diffusiondrive} adopts a diffusion policy, and LTF~\cite{dauner2024navsim} serves as a vision-only baseline. All end-to-end models are evaluated open-loop using official pre-trained weights without fine-tuning.

\subsection{Comparison with Baselines.}

\textbf{Trajectory generation model.} As shown in Table~\ref{tab:traj_compare}, ControlPed achieves superior trajectory realism with $\mathrm{FD}$ of $15.5$. Although Adv-BMT yields the highest $\mathrm{CR}$, it exhibits severe realism degradation and a zero $\mathrm{Span}$, which fails to generate near-miss scenarios. 
Conversely, CCDiff and CounterScene maintain realistic trajectories with moderate criticality, but generate on a narrower trajectory range. ControlPed bridges this gap, attaining the highest $\mathrm{HBR}$ along with a high $\mathrm{CR}$. The margin between $\mathrm{HBR}$ and $\mathrm{CR}$ highlights its ability to synthesize realistic near-miss events that force ego emergency braking without collision.

\begin{table}[!h]
\centering
\caption{\textbf{Critical trajectory generation.} $\dag$ marks values that are numerically extreme but degenerate.}
\label{tab:traj_compare}
\setlength{\tabcolsep}{3pt}
\footnotesize
\resizebox{\columnwidth}{!}{%
\begin{tabular}{lcccccc}
\toprule
\multirow{2}{*}{Method} & Realism & \multicolumn{2}{c}{Criticality} & \multicolumn{3}{c}{Diversity} \\
\cmidrule(lr){2-2}\cmidrule(lr){3-4}\cmidrule(lr){5-7}
 & FD $\downarrow$ & CR $\uparrow$ & HBR $\uparrow$ & Cov $\uparrow$ & $M_{\mathrm{div}}\uparrow$ & Span $\uparrow$ \\
\midrule
Adv-BMT~\cite{liu2026adv}                & 202.7 & \textbf{0.750}$^{\dag}$ & 0.293 & 0.750 & \textbf{12.35}$^{\dag}$ & 0.000 \\
CCDiff~\cite{lin2025causal}              & 63.77 & 0.104 & 0.281 & 0.337 & 0.12 & 0.164 \\
CounterScene~\cite{jing2026counterscene} & 47.92 & 0.113 & 0.280 & 0.304 & 0.09 & 0.131 \\
\midrule
\textbf{ControlPed} & \textbf{15.50} & 0.304 & \textbf{0.735} & \textbf{0.793} & 0.74 & \textbf{0.489} \\
\bottomrule
\end{tabular}%
}
\end{table}

\begin{figure}[!htbp]
    \centering
    \includegraphics[width=\columnwidth]{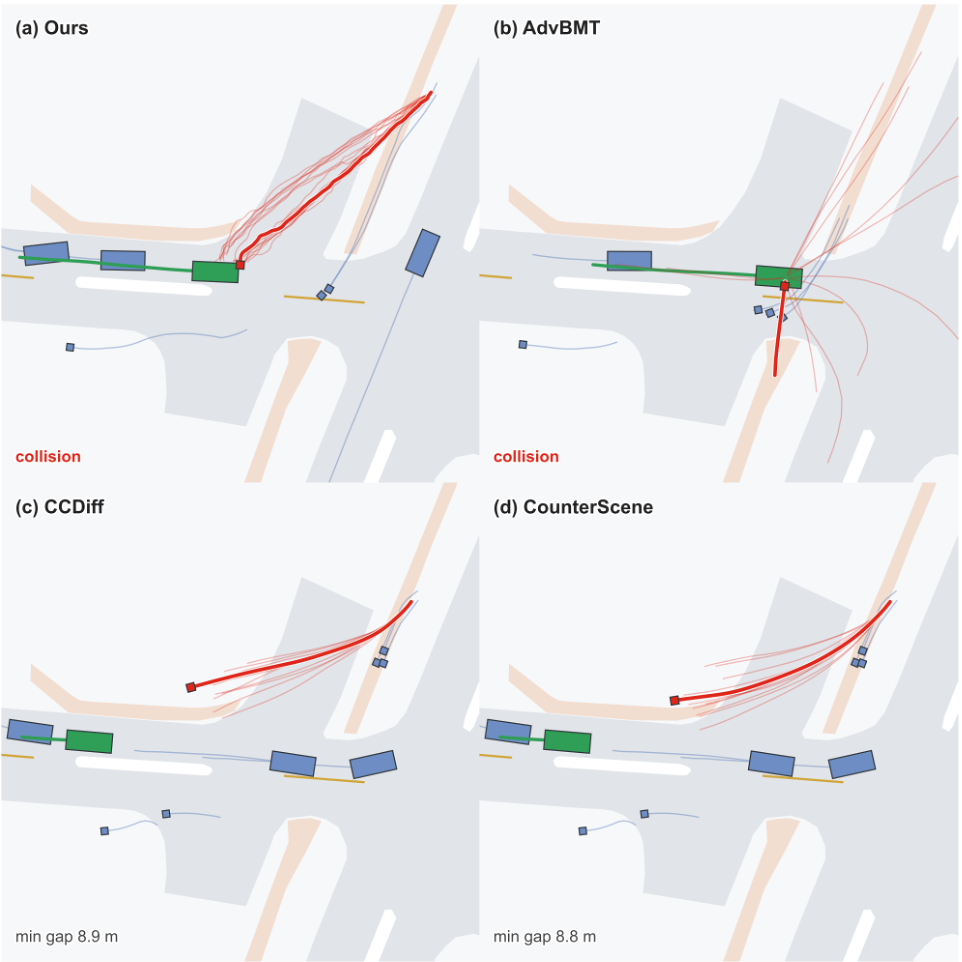}
    \caption{Qualitative comparison of hazardous trajectory generation. (a) ControlPed, (b) Adv-BMT, (c) CCDiff, (d) CounterScene.}
    \label{fig:traj_vis}
    \vspace{-8pt}
\end{figure}

ControlPed achieves superior spatial diversity and coverage, as evidenced by
high $\mathrm{Cov}$ and $\mathrm{Span}$. Qualitatively (Fig.~\ref{fig:traj_vis}),
Adv-BMT produces unconstrained fan-like trajectory scatter, whereas ControlPed
converges into plausible hazard corridors that reach spatial contact.
In contrast, CCDiff and CounterScene remain conservative, terminating at
safety gap of $8.9$\,m and $8.8$\,m, respectively.

\textbf{Ablation study for trajectory generation.} As shown in Table~\ref{tab:traj_ablation}, we evaluate the goal heading $\bm{g}$
and the two-stage training. Removing $\bm{g}$ causes the most
severe degradation, dropping $\mathrm{CR}$ by $85.5\%$ and narrowing
$\mathrm{Span}$ by $46.8\%$, verifying that the criticality is explicitly guided by $\bm{g}$. Omitting Stage-2 fine-tuning preserves headings toward $\bm{g}$ with marginal $\mathrm{Cov}$ drop, but $\mathrm{HBR}$ plummets from $0.735$ to $0.282$ and $\mathrm{FD}$ degrades from $15.50$ to $22.04$. This confirms that conflict fine-tuning simultaneously boosts both criticality and trajectory realism. Removing both components yields the worst performance and fails to generate effective hazards.

\begin{table}[!htbp]
\centering
\caption{\textbf{Ablation on the goal heading and two-stage training.}}
\label{tab:traj_ablation}
\setlength{\tabcolsep}{4.5pt}
\footnotesize
\begin{tabular}{lcccccc}
\toprule
\multirow{2}{*}{Setting} & Realism & \multicolumn{2}{c}{Criticality} & \multicolumn{3}{c}{Diversity} \\
\cmidrule(lr){2-2}\cmidrule(lr){3-4}\cmidrule(lr){5-7}
 & FD $\downarrow$ & CR $\uparrow$ & HBR $\uparrow$ & Cov $\uparrow$ & $M_{\mathrm{div}}\uparrow$ & Span $\uparrow$ \\
\midrule
w/o goal   & 27.67 & 0.044 & 0.330 & 0.391 & 0.420 & 0.260 \\
w/o fine-tune     & 22.04 & 0.171 & 0.282 & 0.696 & 0.675 & 0.386 \\
w/o both   & 37.70 & 0.031 & 0.131 & 0.326 & 0.209 & 0.175 \\
\midrule
\textbf{ControlPed} & \textbf{15.50} & \textbf{0.304} & \textbf{0.735} & \textbf{0.793} & \textbf{0.740} & \textbf{0.489} \\
\bottomrule
\end{tabular}
\end{table}

\textbf{Motion generation model.} As shown in Table~\ref{tab:motion_compare}, ControlPed achieves superior distribution realism with an $\mathrm{FID}$ of $0.517$ and tight heading alignment with an $\mathrm{HED}$ of $0.27^{\circ}$. Although OmniControl matches ControlPed on $\mathrm{MPHE}$ and $\Delta\mathrm{CSR}$, its higher
$\mathrm{FID}$ and $\mathrm{HED}$ reveal unnatural poses with decoupled
locomotion headings. Notably, PriorMDM and MDM outperform on R@3 and $\mathrm{MM\text{-}Dist}$. This demonstrates that generating stereotyped, prototypical motions boosts retrieval metrics at the cost of diversity, a mode collapse phenomenon explicitly captured by extreme $\Delta\mathrm{CSR}$.

\begin{table}[!ht]
\centering
\caption{\textbf{Comparison with baselines on motion generation.}}
\label{tab:motion_compare}
\scriptsize
\setlength{\tabcolsep}{3pt}
\resizebox{\columnwidth}{!}{%
\begin{tabular}{lcccccc}
\toprule
Model & FID $\downarrow$ & R@3 $\uparrow$ & MM-Dist $\downarrow$ & $\Delta$CSR $\rightarrow 0$ & MPHE $\downarrow$ & HED $\downarrow$ \\
\midrule
\textit{Real} & \textit{0.275} & \textit{0.367} & \textit{4.980} & \textit{0} & \textit{0} & \textit{0} \\
\midrule
MDM~\cite{tevet2022human}            & 0.715 & 0.382 & 4.920 & $+$0.356 & 12.030 & 5.960 \\
CondMDI~\cite{cohan2024flexible}     & 2.310 & 0.265 & 5.590 & $-$0.254 & 15.600 & 2.170 \\
StableMoF.~\cite{huang2024stablemofusion} & 1.512 & 0.310 & 5.600 & $+$0.322 & 6.290 & 1.190 \\
PriorMDM~\cite{shafir2024human}      & 1.173 & \textbf{0.389} & \textbf{4.760} & $+$0.421 & 10.640 & 0.930 \\
OmniControl~\cite{xie2024omnicontrol} & 1.145 & 0.312 & 5.650 & \textbf{$-$0.019} & \textbf{6.010} & 0.680 \\
\midrule
\textbf{ControlPed} & \textbf{0.517} & 0.317 & 5.530 & $+$0.054 & 6.810 & \textbf{0.270} \\
\bottomrule
\end{tabular}%
}
\end{table}

\begin{figure*}[t]
    \centering
    \includegraphics[width=\textwidth]{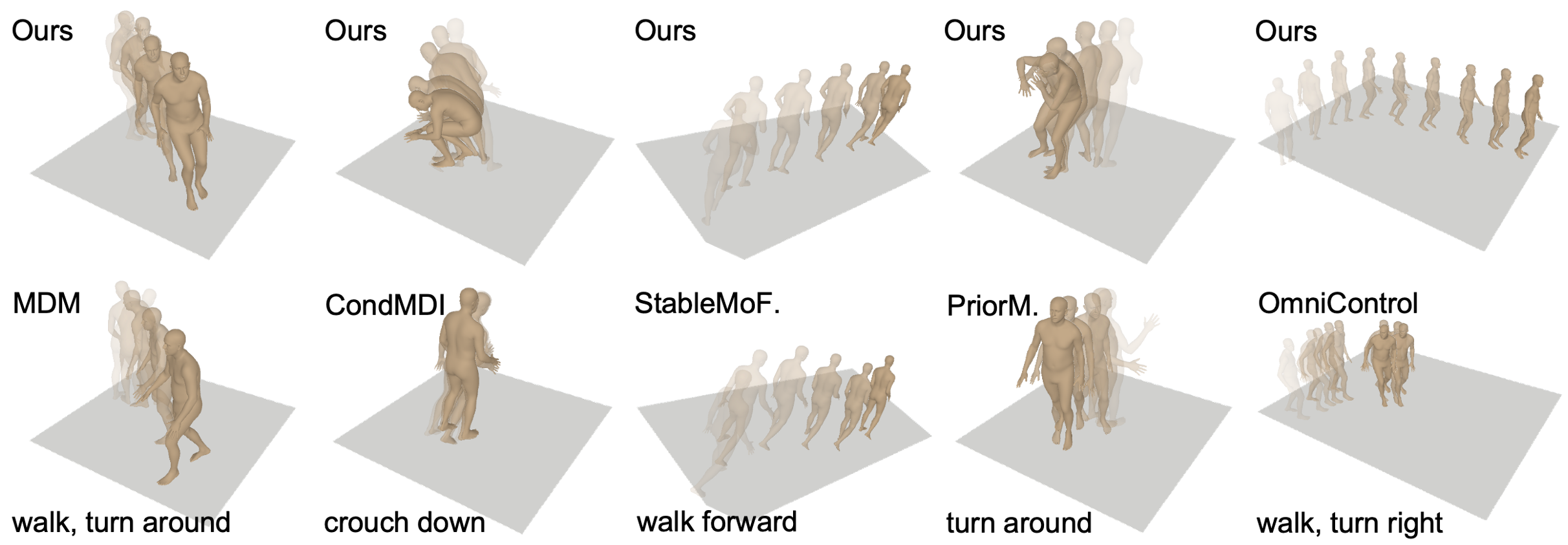}
    \caption{Qualitative comparison of motion generation model. Each column shows ControlPed (top) and a baseline (bottom) under the same text prompt}
    \label{fig:motion_vis}
    \vspace{-15pt}
\end{figure*}

Qualitatively (Fig.~\ref{fig:motion_vis}), baseline methods exhibit distinct
motion artifacts. MDM suffers heading misalignment, CondMDI remains upright
during \textit{crouch down} with only arm gestures, StableMoFusion exhibits
foot sliding, PriorMDM collapses into stationary poses during
\textit{turn around}, and OmniControl produces motion jitter. In contrast,
ControlPed generates plausible and text-aligned human
motions.

\textbf{Ablation study for motion generation.}
As shown in Table~\ref{tab:motion_ablation}, the full ControlPed achieves
optimal performance in $\mathrm{FID}$, $\Delta\mathrm{CSR}$,
$\mathrm{MPHE}$ and $\mathrm{HED}$. The conditioning design has the most
significant impact. Replacing adaptive layer normalization with token
concatenation (w/o adaLN) degrades $\mathrm{FID}$ from $0.517$ to
$1.179$ and $\mathrm{MPHE}$ from $6.81$ to $9.08$\,cm, demonstrating that
per-sublayer modulation prevents control signal dilution in low-pose actions.
Predicting absolute root orientation instead of residual yaw
(w/o reyaw) primarily harms heading alignment, elevating
$\mathrm{HED}$ from $0.27^{\circ}$ to $2.28^{\circ}$. Omitting
geometric losses (w/o geo) severely spikes $\Delta\mathrm{CSR}$ to
$+0.356$, proving their role in regularizing latent distribution. Crucially,
removing category embeddings (w/o cat) boosts R@3 and
MM-Dist while incurring the worst $\mathrm{FID}$ and highest
$\Delta\mathrm{CSR}$. This confirms that losing category conditioning induces
prototype mode collapse, which inflates retrieval scores at the severe expense
of distributional fidelity.

\begin{table}[!h]
\centering
\caption{\textbf{Ablation on model components of ControlPed.}}
\label{tab:motion_ablation}
\scriptsize
\setlength{\tabcolsep}{4pt}
\resizebox{\columnwidth}{!}{%
\begin{tabular}{lcccccc}
\toprule
Setting & FID $\downarrow$ & R@3 $\uparrow$ & MM-Dist $\downarrow$ & $\Delta$CSR $\rightarrow 0$ & MPHE $\downarrow$ & HED $\downarrow$ \\
\midrule
w/o adaLN & 1.179 & 0.321 & 4.730 & $+$0.327 & 9.080 & 1.730 \\
w/o reyaw & 1.076 & 0.289 & 5.450 & $+$0.228 & 7.240 & 2.280 \\
w/o geo   & 0.660 & 0.327 & 5.060 & $+$0.356 & 7.000 & 0.980 \\
w/o cat   & 1.260 & \textbf{0.376} & \textbf{4.640} & $+$0.607 & 7.250 & 0.790 \\
\midrule
\textbf{ControlPed} & \textbf{0.517} & 0.317 & 5.530 & \textbf{$+$0.054} & \textbf{6.810} & \textbf{0.270} \\
\bottomrule
\end{tabular}%
}
\end{table}

\subsection{End-to-End Safety Evaluation}
Evaluating seven end-to-end driving models under rendered pedestrian
hazards reveals a severe average $\mathrm{HDScore}$ drop from $88.8$ to $47.4$
(Table~\ref{tab:e2e}). Performance degradation is highly metric-specific.
The drivable area compliance ($\mathrm{DAC}$) remains largely unaffected, showing
that planners preserve drivable space awareness. Instead, the score collapse is
primarily driven by sharp declines in route completion ($\mathrm{RC}$),
no collision ($\mathrm{NC}$), and time-to-collision ($\mathrm{TTC}$).
This indicates that models do not lose structural road discipline in
pedestrian conflicts. Rather, they fail to resolve critical interaction
dynamics, causing collisions that prematurely terminate route execution.

\begin{table}[!ht]
\centering
\caption{\textbf{End-to-end driving evaluation.}
Reconstructed original scene (left) and safety-critical scenario (right).}
\label{tab:e2e}
\scriptsize
\setlength{\tabcolsep}{2pt}
\renewcommand{\arraystretch}{1.15}
\resizebox{\columnwidth}{!}{%
\begin{tabular}{lcccccc}
\toprule
Model & RC $\uparrow$ & NC $\uparrow$ & DAC $\uparrow$ & TTC $\uparrow$ & COM $\uparrow$ & HDScore $\uparrow$ \\
\midrule
DiffDrive~\cite{liao2025diffusiondrive} & 99.0/62.8 & \textbf{99.0}/86.2 & \textbf{99.9}/99.8 & \textbf{94.8}/84.9 & 99.0/99.4 & 94.6/51.9 \\
SSR~\cite{li2025navigation} & \textbf{100.0}/62.7 & 98.6/84.6 & 99.6/99.6 & 94.4/84.0 & 99.5/99.4 & \textbf{95.1}/51.1 \\
VAD~\cite{jiang2023vad} & 99.6/63.7 & 97.6/84.3 & 99.4/99.1 & 92.1/83.4 & 99.4/99.9 & 92.7/50.8 \\
LAW~\cite{li2025enhancing} & \textbf{100.0}/62.8 & 97.7/84.2 & 98.7/98.4 & 92.6/82.5 & \textbf{99.9}/100.0 & 92.8/49.7 \\
LTF~\cite{dauner2024navsim} & 98.8/60.7 & 96.7/84.4 & 99.0/98.9 & 93.9/83.7 & 90.3/89.9 & 90.0/47.7 \\
UniAD~\cite{hu2023planning} & 80.7/47.2 & 90.1/79.2 & 90.8/90.7 & 79.9/72.6 & 87.2/83.7 & 62.4/30.2 \\
SparseDrive~\cite{sun2025sparsedrive} & \textbf{100.0}/62.2 & 98.5/85.2 & 99.3/99.2 & 93.8/83.6 & 99.5/99.9 & 94.3/50.6 \\
\midrule
\textit{Mean drop} & $-$36.6 & $-$12.9 & $-$0.2 & $-$9.5 & $-$0.4 & $-$41.4 \\
\bottomrule
\end{tabular}%
}
\end{table}

\begin{figure}[H]
    \vspace{4pt}
    \centering
    \includegraphics[width=\columnwidth]{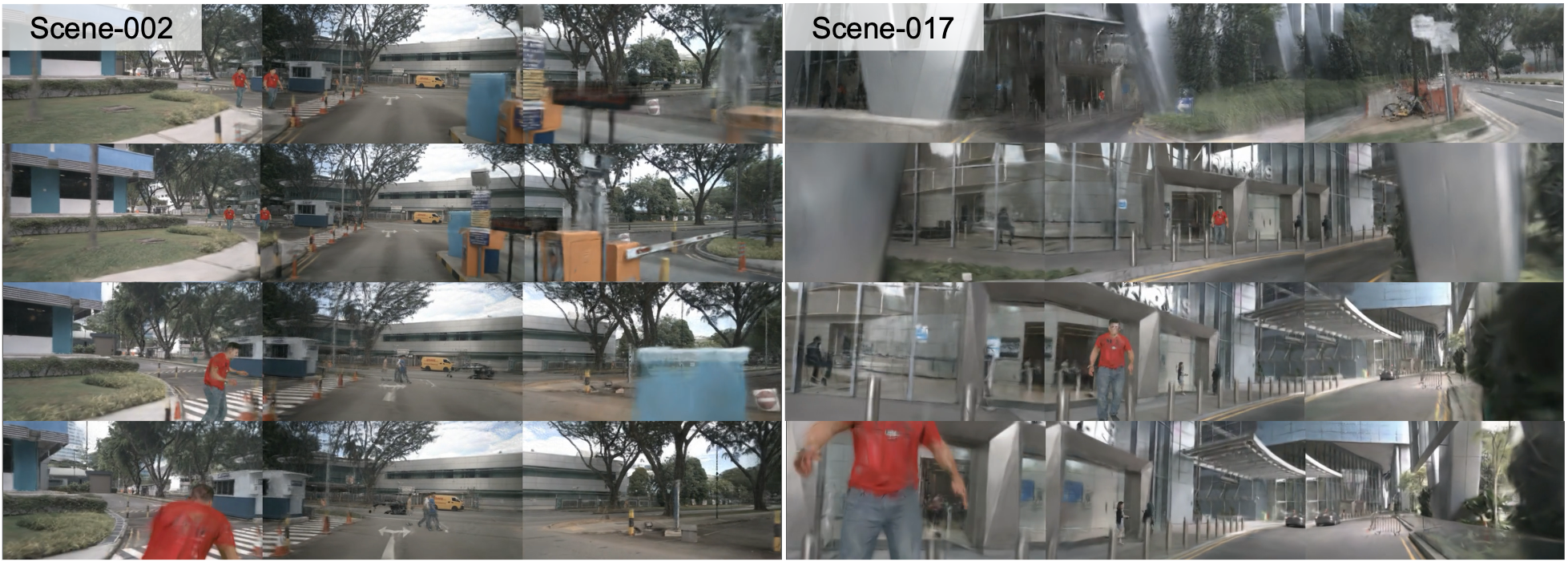}
    \vspace{-10pt}
    \caption{Qualitative rendered safety-critical scenarios at a representative keyframe. Pedestrians maintain multi-view spatial consistency.}
    \label{fig:render_vis}
    \vspace{4pt}
\end{figure}

As illustrated in Fig.~\ref{fig:render_vis}, rendered hazardous pedestrians maintain robust multi-view temporal consistency across all surround cameras, providing sensor-level photorealism that drives end-to-end perception and planning pipelines.

\section{CONCLUSIONS}
We propose ControlPed, a controllable and photorealistic scenario generation method built upon our HazardPed dataset. This work provides a benchmark for regulatory agencies and third-party institutions to validate end-to-end autonomous driving safety prior to real-world deployment. Although ControlPed currently operates in an open-loop setup, extending it to closed-loop interactive simulation represents a fruitful avenue for future research.

\addtolength{\textheight}{-12cm}   





\bibliographystyle{IEEEtran}
\bibliography{references}

\end{document}